\documentclass{article}
\usepackage{amsmath}
\usepackage{paralist}
\usepackage{graphics}
\usepackage{epsfig}
\usepackage{graphicx}
\usepackage{epstopdf}
\usepackage{caption}
\usepackage[labelformat=simple]{subcaption}

\usepackage[colorlinks=true]{hyperref}
\usepackage{todonotes}
\usepackage{graphbox} 
\usepackage{pgffor}
\hypersetup{urlcolor=blue, citecolor=green}
\usepackage{xcolor}
\usepackage{tikz}
\usepackage{varwidth}
\usepackage{tcolorbox}
\usepackage{mathtools} 
\usepackage{tablefootnote} 

\newcommand{\norm}[1]{\left\lVert#1\right\rVert}
\newcommand{\frob}[1]{\norm{#1}_\text{F}}
\newcommand{\ourtestdataset}{our test dataset}
\newcommand{\hdctestdataset}{official challenge test dataset}
\newcommand{\arXiv}[1]{\texttt{\href{http://arxiv.org/abs/#1}{arXiv:#1}}}

\title{Deblurring Photographs of Characters Using Deep Neural Networks} 

\author{Thomas Germer\footnote{Corresponding author.}\\
Heinrich Heine University Düsseldorf\\
\small{\texttt{thomas.germer@hhu.de}}
\and
Tobias Uelwer\footnote{Part of this work was done at the Heinrich Heine University Düsseldorf.}\\
Technical University Dortmund\\
\small{\texttt{tobias.uelwer@tu-dortmund.de}}
\and
Stefan Harmeling\textsuperscript{\textdagger}\\
Technical University Dortmund\\
\small{\texttt{stefan.harmeling@tu-dortmund.de}}}

\begin{document}

\date{} 
\maketitle

\begin{abstract}
  In this paper, we present our approach for the Helsinki Deblur
  Challenge (HDC2021). The task of this challenge is to deblur images
  of characters without knowing the point spread function (PSF). The
  organizers provided a dataset of pairs of sharp and blurred
  images. Our method consists of three steps: First, we estimate a
  warping transformation of the images to align the sharp images with
  the blurred ones. Next, we estimate the PSF using a
  quasi-Newton method. The estimated PSF allows to generate additional
  pairs of sharp and blurred images. Finally, we train a deep convolutional
  neural network to reconstruct the sharp images from the blurred
  images. Our method is able to successfully reconstruct images from
  the first 10 stages of the HDC 2021 dataset. Our code is available at \url{https://github.com/hhu-machine-learning/hdc2021-psfnn}.
\end{abstract}

\section{Introduction}

Blurring a sharp image $G^S$ can be modeled as a convolution with a point-spread-function $P$
\begin{equation}
	G^B = P * G^S  + \varepsilon,
\end{equation}
where $*$ denotes the convolution operation and $\varepsilon$ is an unknown noise term.
The process of reconstructing the sharp image $G^S$ given the blurred images $G^B$ is referred to as deblurring. Deblurring problems can be categorized into two classes: if the PSF  $P$ is known, the problem is a non-blind deblurring problem, whereas if $P$ is unknown, it is a blind deblurring problem.

\subsection{Task description}

\begin{figure}
    \centering
    \includegraphics[width=0.8\linewidth]{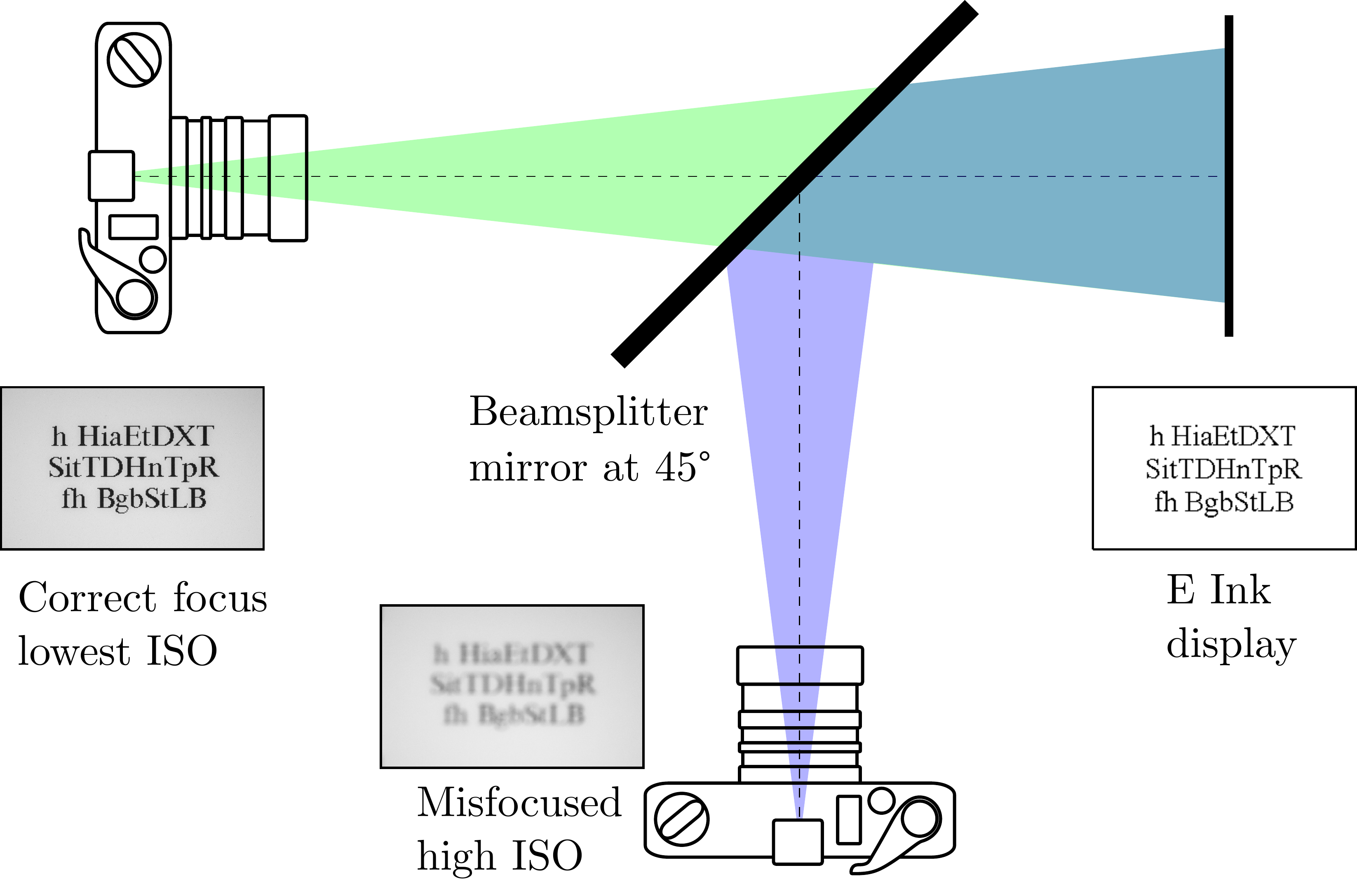}

    \caption{Simplified experimental setup reproduced from the HDC2021 description of photographic data \cite{juvonen2021helsinki}, consisting of two cameras and a beamsplitter mirror that allows both cameras to capture images of the E Ink display. One camera is correctly focused with low ISO setting while the other camera is misfocused with high ISO setting, resulting in noisy and blurry images.}
    \label{fig:setup}
\end{figure}

\begin{figure}
	\centering
	\setlength{\tabcolsep}{5pt}
	\newcommand{\imgwidth}{2.9cm}
	\renewcommand{\arraystretch}{2}
	\begin{tabular}{rlll} 
		Stage 1 & \includegraphics[width=\imgwidth, align=c]{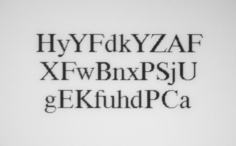} & \includegraphics[width=\imgwidth, align=c]{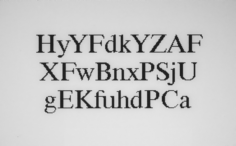} \vspace{0.1cm} & \includegraphics[width=\imgwidth, align=c]{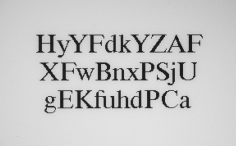} \vspace{0.1cm} \\
		Stage 2 & \includegraphics[width=\imgwidth, align=c]{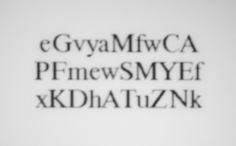} & \includegraphics[width=\imgwidth, align=c]{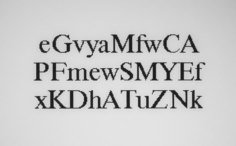} \vspace{0.1cm} & \includegraphics[width=\imgwidth, align=c]{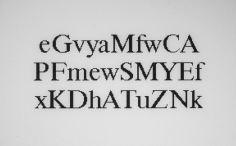} \vspace{0.1cm} \\
		Stage 3 & \includegraphics[width=\imgwidth, align=c]{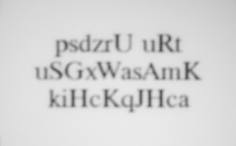} & \includegraphics[width=\imgwidth, align=c]{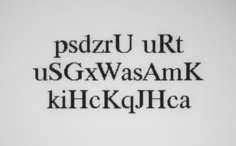} \vspace{0.1cm} & \includegraphics[width=\imgwidth, align=c]{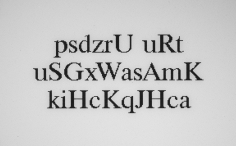} \vspace{0.1cm} \\
		Stage 4 & \includegraphics[width=\imgwidth, align=c]{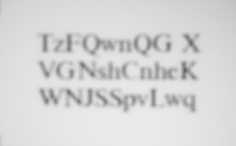} & \includegraphics[width=\imgwidth, align=c]{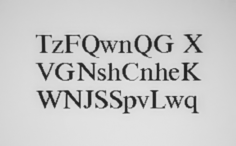} \vspace{0.1cm} & \includegraphics[width=\imgwidth, align=c]{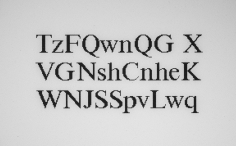} \vspace{0.1cm} \\
		Stage 5 & \includegraphics[width=\imgwidth, align=c]{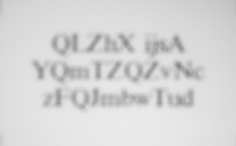} & \includegraphics[width=\imgwidth, align=c]{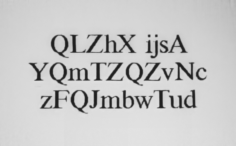} \vspace{0.1cm} & \includegraphics[width=\imgwidth, align=c]{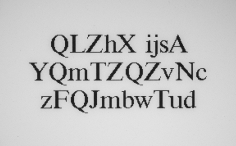} \vspace{0.1cm} \\
		Stage 6 & \includegraphics[width=\imgwidth, align=c]{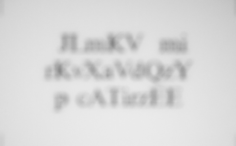} & \includegraphics[width=\imgwidth, align=c]{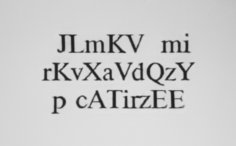} \vspace{0.1cm} & \includegraphics[width=\imgwidth, align=c]{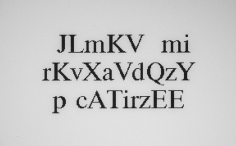} \vspace{0.1cm} \\
		Stage 7 & \includegraphics[width=\imgwidth, align=c]{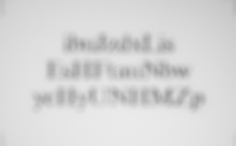} & \includegraphics[width=\imgwidth, align=c]{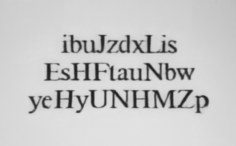} \vspace{0.1cm} & \includegraphics[width=\imgwidth, align=c]{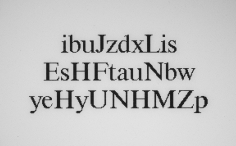} \vspace{0.1cm} \\
		Stage 8 & \includegraphics[width=\imgwidth, align=c]{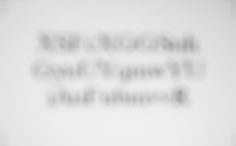} & \includegraphics[width=\imgwidth, align=c]{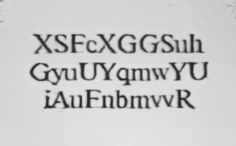} \vspace{0.1cm} & \includegraphics[width=\imgwidth, align=c]{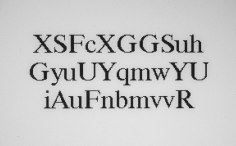} \vspace{0.1cm} \\
		Stage 9 & \includegraphics[width=\imgwidth, align=c]{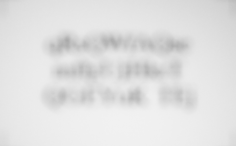} & \includegraphics[width=\imgwidth, align=c]{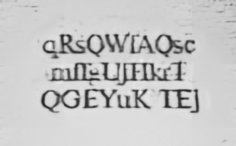} \vspace{0.1cm} & \includegraphics[width=\imgwidth, align=c]{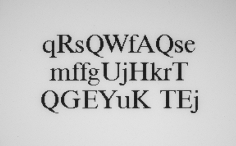} \vspace{0.1cm} \\
		Stage 10 & \includegraphics[width=\imgwidth, align=c]{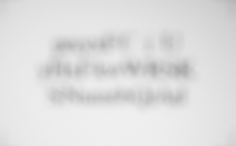} & \includegraphics[width=\imgwidth, align=c]{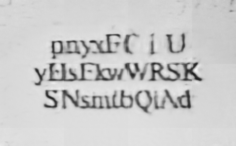} \vspace{0.1cm} & \includegraphics[width=\imgwidth, align=c]{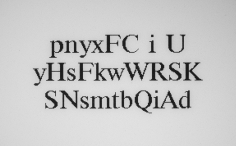} \vspace{0.1cm} \\
	\end{tabular}
	\caption{Blurry images, deblurred images and sharp images.}\label{fig:original-deblurred}
  \end{figure}

The dataset of the Helsinki Deblur
Challenge~\cite{juvonen2021helsinki} consists of pairs of sharp and
blurry monochrome photos of several texts displayed on an E Ink screen. Example images are shown in Figure~\ref{fig:original-deblurred}. The pictures are taken using a setup consisting of two cameras and a beam splitter mirror that allows both cameras to take a picture of the same E Ink display. The first camera is configured to shoot sharp images with low ISO setting while the second camera is out of focus with high ISO setting. The latter camera produces blurry and noisy images. The setup is shown in Figure~\ref{fig:setup}. 
The images are subdivided into $20$ stages, where the images of each stage exhibit a different level of blur. Each stage consists of $200$ images of size $2360 \times 1460$ pixels showing three lines of text in one of two font styles, $100$ images per font. For each of the images, a text file is provided that contains the characters shown on that image.

The performance of a deblurring algorithm is measured by the
Levenshtein distance \cite{levenshtein1966binary} between the ground truth text
and the text obtained via optical character recognition (OCR) from the
deblurred image using Tesseract~\cite{smith2007overview}. However,
tuning the method solely for deblurred text can have the disadvantage
that the reconstructions resemble text also for natural images.  Thus,
the algorithm must pass an additional test where the deblurring capabilities
are evaluated on natural images shown on the E Ink display.

\subsection{Related work}

Early work applying machine learning to the non-blind deblurring
problem include Schuler et al.~\cite{schuler2013machine}, who use a neural network to remove artifacts after direct deconvolution. Later, Xu et al.~\cite{xu2014deep} and Schuler et al.~\cite{schuler2015learning} apply neural networks to the blind deblurring problem, where the PSF is not given.

Kupyn et al. \cite{kupyn2018deblurgan,kupyn2019deblurgan} use a conditional generative adversarial network to approach the deblurring
problem. Non-uniform motion deblurring, i.e. the problem of
deblurring an image where the PSF differs at each location of the
image, leveraging deep neural networks has been analyzed by Sun et
al.~\cite{sun2015learning}. Nah et al.~\cite{nah2017deep} propose a
multi-stage approach to solve blind debluring problems.  Recently, Zamir
et al.~\cite{zamir2021multi} also discuss a multi-stage approach
involving attention mechanisms for deblurring.

\section{Our approach}

With the recent successes of neural network-based methods on several
image processing
tasks~\cite{kim2016accurate,zamir2021multi,zhang2018ffdnet}, adapting
such an approach to this task seemed like a promising direction to
take.  However, a neural network trained only on text images will not
perform well when applied to natural images.  In other words, it
will not generalize to other image modalities.  Therefore, it will be
useful to generate blurry natural images (non-text), that have the
same blur as the images obtained from the experimental setup in
Figure~\ref{fig:setup}.

The general outline of our approach is thus:
\begin{enumerate}
	\item Estimate a forward model to simulate the blurring process.
	\item Apply the forward model to a dataset of sharp natural images to obtain blurry natural images.
	\item Train a neural network on both blurry natural images and blurry text photos to produce sharp images.
\end{enumerate}
The forward model can be further subdivided into two steps. First, we warp the sharp input images so they are more closely aligned to the blurry output images. This facilitates easier learning for a convolutional neural network later on, since it does not have to learn a translation task first. In addition, the receptive field of the neural network can be smaller, since the information required for deblurring will be more localized. The warping operator has to be estimated prior to this step. Second, we perform a blurring operation employing a PSF, which also has to be estimated beforehand.

\subsection{Warping matrix estimation}

Since the exact warping operation is unknown, we opted for a simple
weighted combination of a yet to be determined $2\times 10$ weight matrix $W$ with third degree polynomials depending on the input pixel coordinates $i$ and $j$ to obtain transformed output coordinates $i'$ and $j'$,
\begin{equation}
	\label{eq:warping}
	 \left[i^\prime, j^\prime \right]^T=W \left[1, i, j, i^2, ij, j^2, i^3, i^2j, ij^2, j^3\right]^T\hspace{-8pt}.
\end{equation}
Furthermore, we use bicubic interpolation, represented by the operator $\Psi$, to obtain a warped image $G_\text{warped}$,
\begin{equation}
	G_\text{warped} = \Psi \left( G, W \right),
\end{equation}
 by evaluating the image $G$ at the new positions $\left[i^\prime, j^\prime \right]$ given an input image $G$ and the weight matrix $W.$
In addition, we define the centering operator $C$ of an image $G$ with width $n$ and height $m$. A pixel $C(G)_{ij}$ of the centered image is the difference between the uncentered pixel $G_{ij}$ minus the average color value $\frac{1}{m n} \sum_{i=1}^m \sum_{j=1}^n G_{ij}$ of the image,
\begin{equation}
	C(G)_{ij} = G_{ij} - \frac{1}{m n} \sum_{i=1}^m \sum_{j=1}^n G_{ij}.
\end{equation}
Centering is helpful for later steps to reduce the difference in brightness between two images.
\begin{figure}
	\centering
	\begin{subfigure}[t]{0.49\textwidth}
		\includegraphics[width=\linewidth]{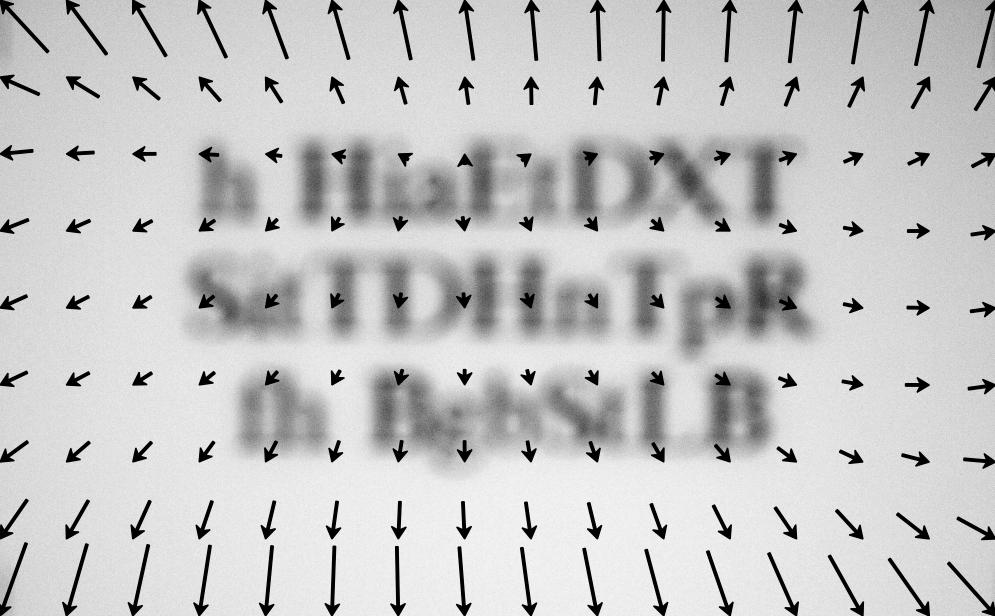}
		\caption{Warping directions superimposed over blurry image.}
		\label{figure:vectors}
	\end{subfigure}
	\hfill
	\begin{subfigure}[t]{0.49\textwidth}
		\includegraphics[width=\linewidth]{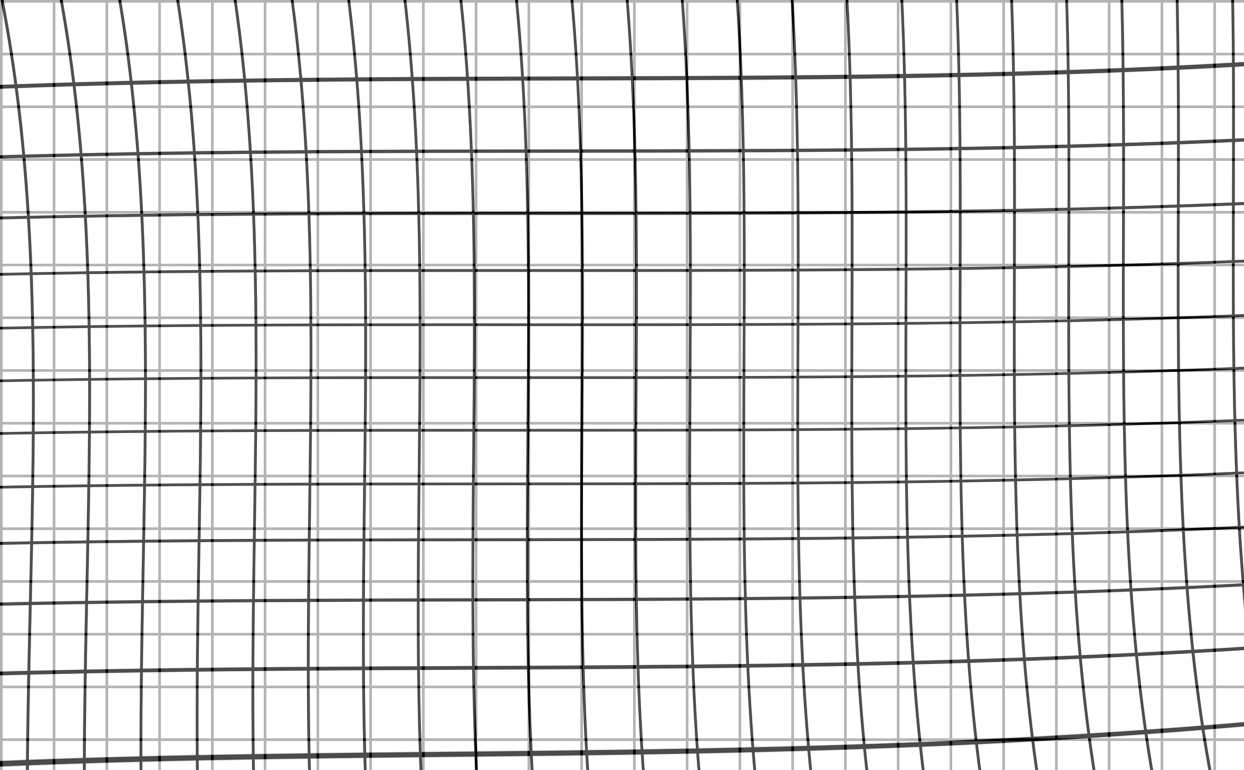}
		\caption{A regular grid before warping (light gray) and after
		warping (black).}
		\label{figure:grid}
		\end{subfigure}
	\caption{Visualization of warping operation.}\label{dewarp}
\end{figure}
To obtain the parameters of the warping transformation, we optimize the loss function $\mathcal{L}_\text{warping}$ to minimize the mean squared error (MSE) between the centered warped blurry image and its corresponding centered sharp image
\begin{equation}
	\label{eq:Lwarping}
	\mathcal{L}_\text{warping}(W) = \frac{1}{m n} \frob{ C\left(\Psi \left( {G^B}, W \right) \right) - C\left(G^S\right) }^2,
\end{equation}
where $\frob{A} \coloneqq \sqrt{ \sum_i \sum_j | a_{ij} |^2 } $ denotes the Frobenius norm.

\begin{figure}
	\centering
	\begin{subfigure}[t]{0.49\textwidth}
		\includegraphics[width=\linewidth]{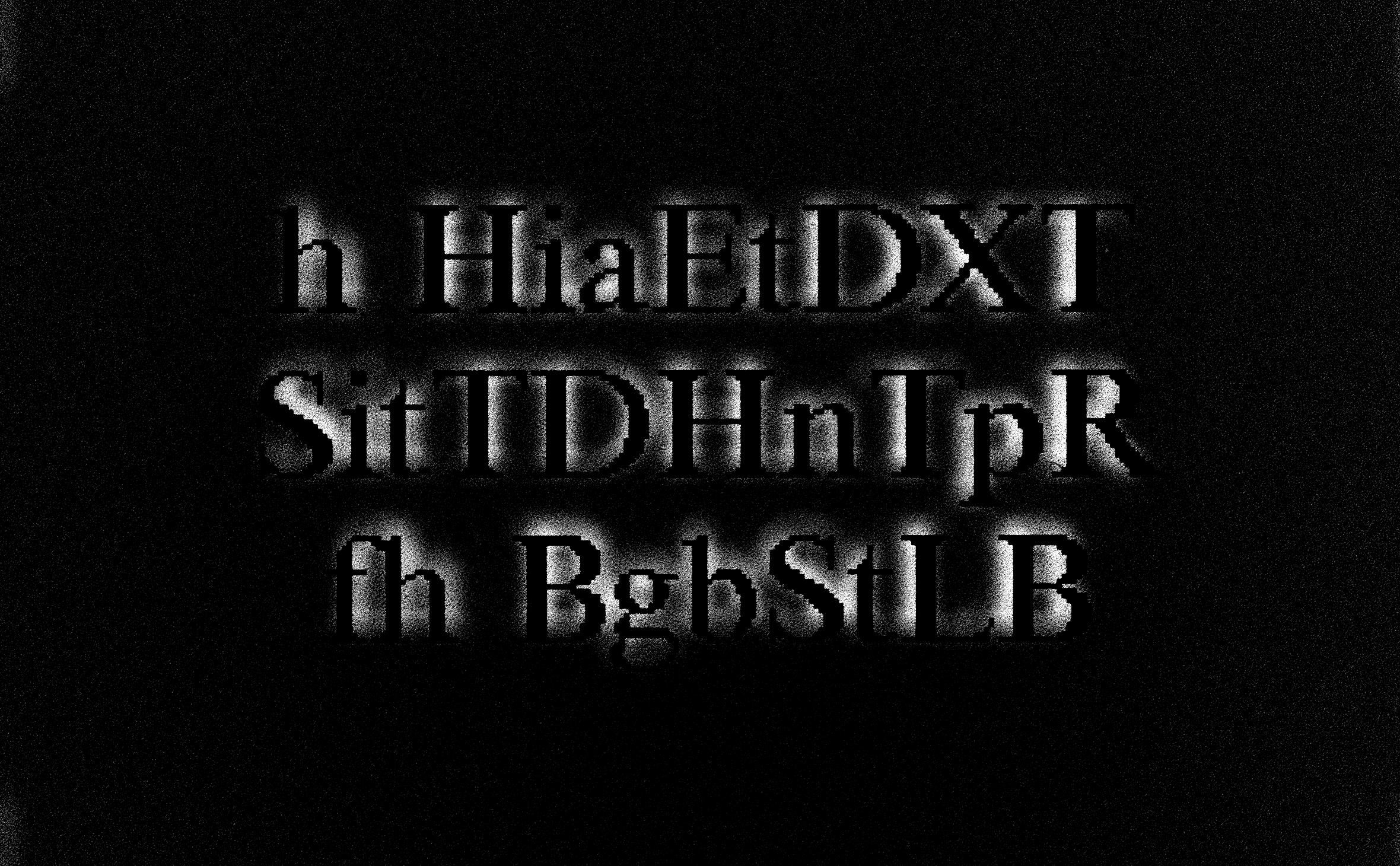}
		\caption{Difference without warping. The characters are not well-aligned.}
		
	\end{subfigure}
	\hfill
	\begin{subfigure}[t]{0.49\textwidth}
		\includegraphics[width=\linewidth]{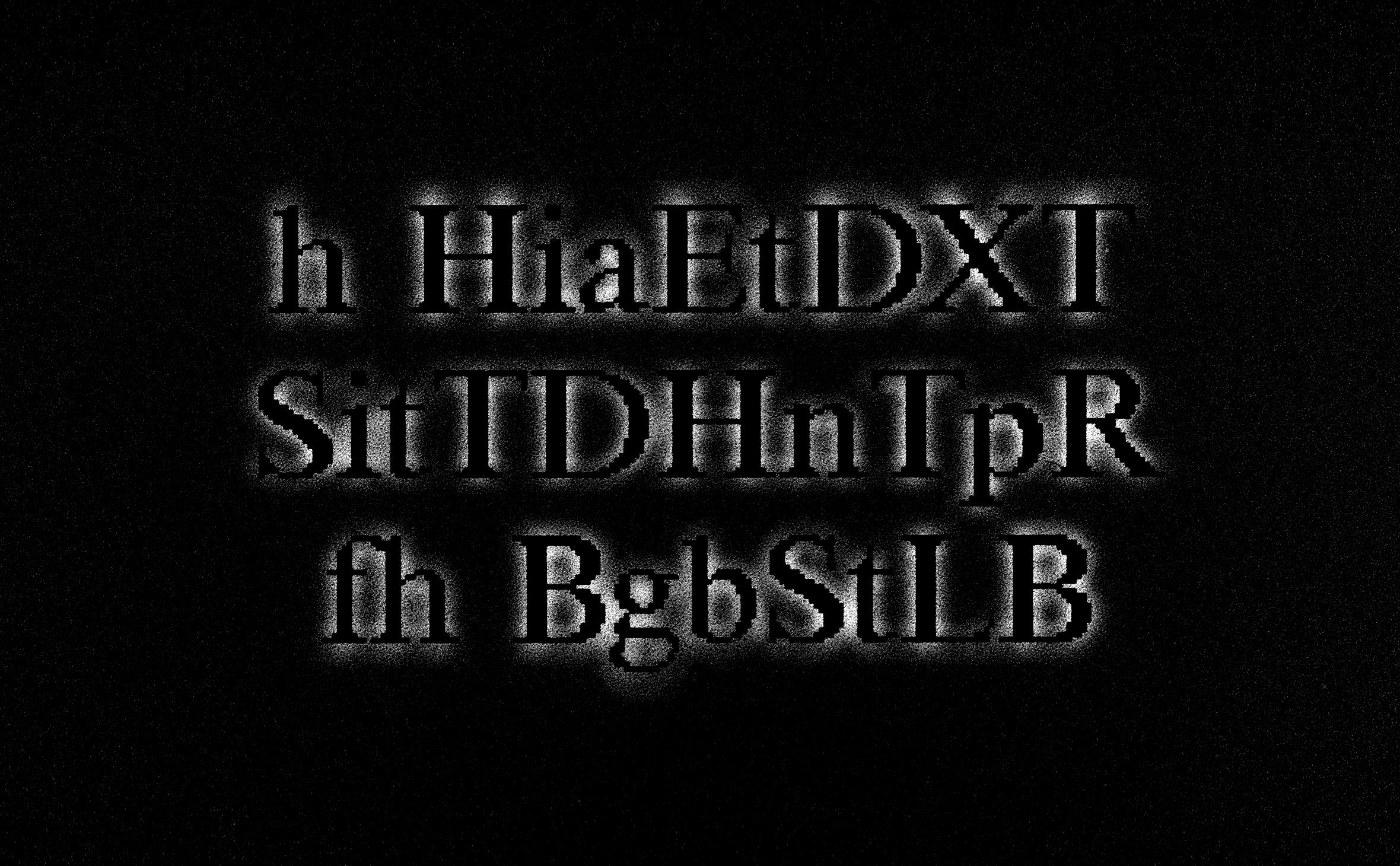}
		\caption{Difference with warping. The characters are well aligned.}
		\label{figure:2}
		\end{subfigure}
	\caption{Difference between sharp and (warped) blurry image.}
	\label{fig:dewarpdifference}
\end{figure}

As can be seen in Figure~\ref{fig:dewarpdifference}, the difference between the sharp and warped blurry image are reduced.
When a regular grid is transformed with the estimated warping operator (Figure~\ref{figure:grid}), it becomes apparent that the warping at the edges of the image is stronger than might be expected. This can be explained by the lack of characters or other distinctive features in that region. Since there are no characters to deblur, this issue has no consequences for the further steps.
In order to allow for a simpler estimation of the PSF, the dataset also contains photos of a vertical line, a horizontal line and a dot. However, we opted to ignore those images and estimate the PSF
directly from the text photos instead, since the text covers a larger
region of the image and therefore conveys more information about the
PSF.

\subsection*{Why degree 3?}

\begin{figure}
	\centering
	\includegraphics[width=\linewidth]{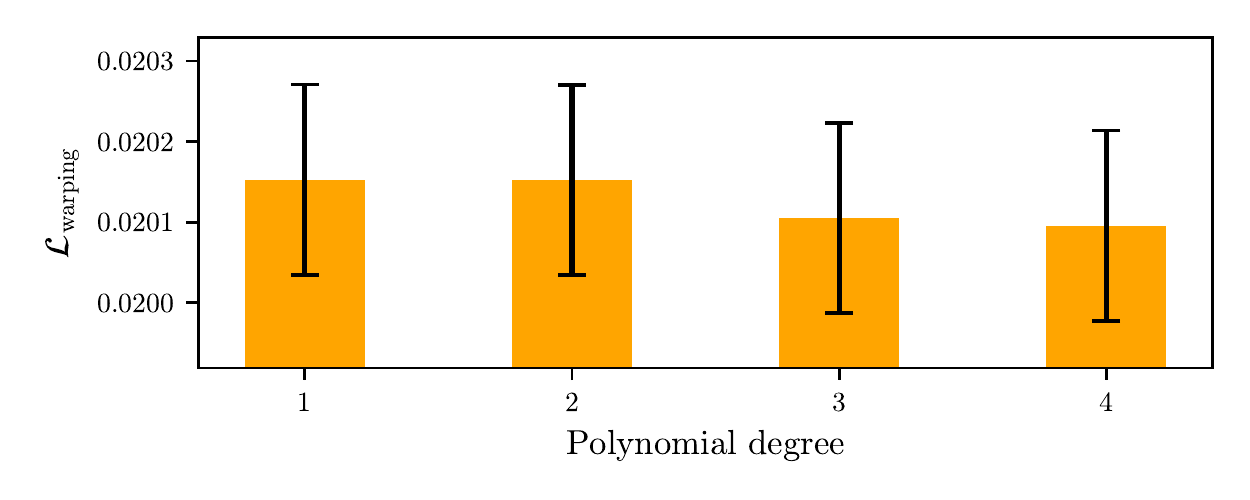}
	
	\caption{Degree of polynomial features (Equation~\ref{eq:warping}) versus $\mathcal{L}_\text{warping}$ (Equation~\ref{eq:Lwarping}) averaged over the image pairs of the fifth stage of the HDC2021 dataset, including standard error bars. Note that the y-axis is offset to emphasize the small difference between losses, which is relatively small compared to the standard error.}
	\label{fig:Lwarping}
\end{figure}

One might consider a higher degree for the polynomial features (Equation~\ref{eq:warping}) to allow for more flexible warping transformations. When considering $\mathcal{L}_\text{warping}$ (Figure~\ref{fig:Lwarping}), we find that the step from degree $2$ to degree $3$ results in the largest error reduction. Even higher degrees yield diminishing returns, especially when taking the comparatively high standard error into account, which is why we settled on degree $3$.

\subsection{Estimation of the point spread function}
To estimate the $p \times p$ point spread function $P$ from a sharp image $G^S$ and a warped blurry image $G^{B}$, we consider minimizing the loss function
\begin{equation}
	\mathcal{L}_\text{PSF}(P, \tau, W) = \frac{1}{m n} \frob{ P * G^S + \tau - \Psi\left(G^B, W\right) }^2 + \lambda \frac{1}{p^2} \sum_{i=1}^p \sum_{j=1}^p \lvert P_{ij} \rvert,
\end{equation}
where $\lambda>0$ is a positive regularization factor to encourage a well-behaved PSF. In addition, we also estimate an additive offset $\tau$ to allow for brightness variation between the two images and refine the warping coefficients $W$ further.
Regularizing the PSF with an $\ell_1$-loss instead of an $\ell_2$-loss reduces noise in the point spread function, but makes it harder to optimize from a numerical point of view. Furthermore, one might consider skipping the step of estimating $W$ and optimize $P$, $\tau$ and $W$ all at once, but in practice, $P$ will converge before $W$ and the optimization will be stuck in a local minima which is difficult to escape.
Nevertheless, with a good initial value for $W,$ a solution can be obtained quickly with PyTorch's implementation~\cite{paszke2019pytorch} of the LBFGS optimizer \cite{liu1989lbfgs}, given that the following performance issue is considered.
Especially, for more challenging stages of the competition, a large PSF is required to model the blurring operation. The computational complexity of a naive convolution implementation scales with $O(m n p^2)$, i.e. with the product of the total number of pixels $m n$ in the image and the squared side length $p^2$ of the PSF. This necessitates a more sophisticated approach, which can be found in a convolution based on the Fast Fourier Transform (FFT). This approach only scales with a more reasonable computational complexity of $O(m n \log n + n m \log m + p^2 \log p)$.

\begin{figure}
\centering
\newcommand{\psfwidth}{0.133 \linewidth}
\begin{tabular}{cccccc}
\includegraphics[width=\psfwidth]{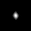} &
\includegraphics[width=\psfwidth]{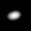} &
\includegraphics[width=\psfwidth]{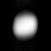} &
\includegraphics[width=\psfwidth]{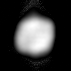} &
\includegraphics[width=\psfwidth]{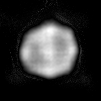} &
\includegraphics[width=\psfwidth]{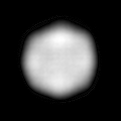}\\
$31\times 31$ & $31\times 31$ & $51\times 51$ & $71\times 71$ & $101\times 101$ & $121\times 121$ \vspace{3mm}\\
\includegraphics[width=\psfwidth]{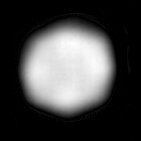} &
\includegraphics[width=\psfwidth]{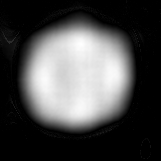} &
\includegraphics[width=\psfwidth]{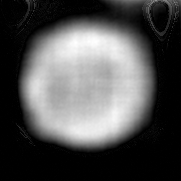} &
\includegraphics[width=\psfwidth]{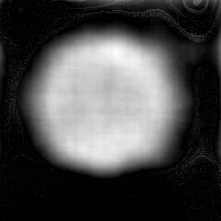} &
\includegraphics[width=\psfwidth]{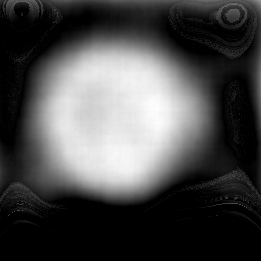} &
\\
$141\times 141$ & $161\times 161$  & $181\times 181$ & $221\times 221$ & $261\times 261$ \\
\end{tabular}
\caption{Estimated point spread functions for stages 0 through 10. For better readability, all PSFs have been scaled to the same image size. They vary between $31 \times 31$ from the smallest size up to $261 \times 261$ for stage 10.}
\label{fig:psfs}
\end{figure}

When looking at the PSFs (Figure~\ref{fig:psfs}) closely, especially for the medium stages, one can recognize an octagonal shape, which coincides with the eight aperture blades of the Canon EF 100mm f/2.8 USM Macro lens used to take the photos \cite{juvonen2021helsinki}. For later stages, the PSF becomes less recognizable, suggesting that the model capabilities have reached their limit.

\subsection{Training dataset}

\begin{figure}
	\centering
	\includegraphics[width=\linewidth]{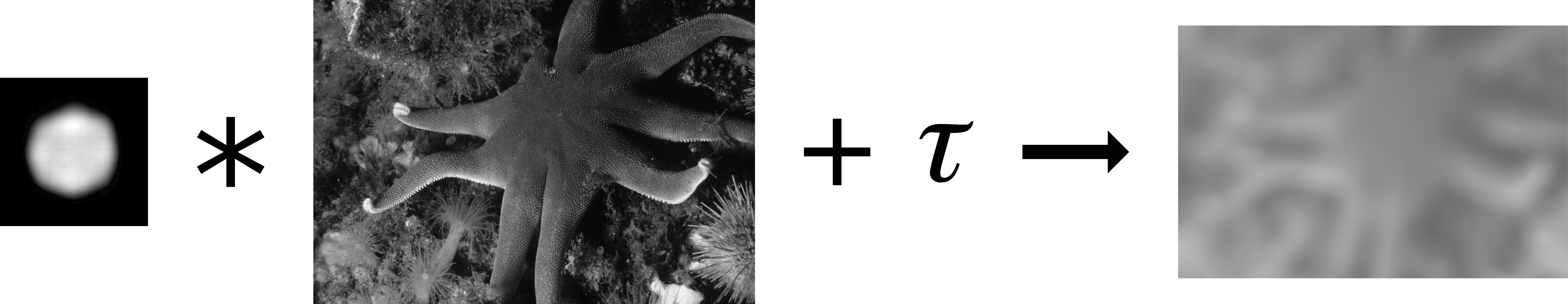}
	
	\caption{The sharp images from the DIV2K dataset $V^S$ (middle left) are convolved with the point spread function $P$ (left) and brightness adjusted ($\tau$) to form a blurry image $V^B$ for training. Note that the blurry image is slightly smaller than the initial sharp image since we only train on the valid convolution region. Before training, we crop the sharp image to the same size.}
	\label{fig:blur_div2k}
\end{figure}

With the forward model completed, we can generate the datasets $D_s$ to train a neural network for deblurring of stage $s$. The datasets contains two sets of image pairs:

\begin{enumerate}
	\item The first 90 sharp images $G^S$ of the HDC2021 dataset as well as the corresponding warped blurry images $\Psi(G^B, W)$ form the first set of image pairs. We keep the remaining 10 images as \ourtestdataset{}, which is not to be confused with the \hdctestdataset{}.
	Since the pixels in the image pair $(G^S, \Psi(G^B, W))$ are mostly aligned due to warping, a neural network can focus entirely on deblurring the images instead of having to spend capacity on pixel alignment.
	\item The second set of images consists of 500 pairs of sharp natural images $V^S$ from the DIV2K dataset~\cite{agustsson2017ntire} as well as their blurry versions $V^B = P * V^S + \tau$ (Figure~\ref{fig:blur_div2k}), which are the sharp images convolved with the PSF $P$ and brightened by the offset $\tau$ estimated from an image of the corresponding stage of the HDC2021 dataset.
	The goal of enriching the dataset with natural images is that a neural network has to perform a more general deblurring operation, which hopefully leads to better generalization, as the HDC2021 challenge required that all methods must be general purpose methods, i.e. they must work for other images than just text images.
\end{enumerate}

\subsection{Deblurring model}
\begin{figure}[htp]
    \centering
    \begin{tikzpicture}
    \draw (0, 0) node[inner sep=0] {\includegraphics[width=\linewidth]{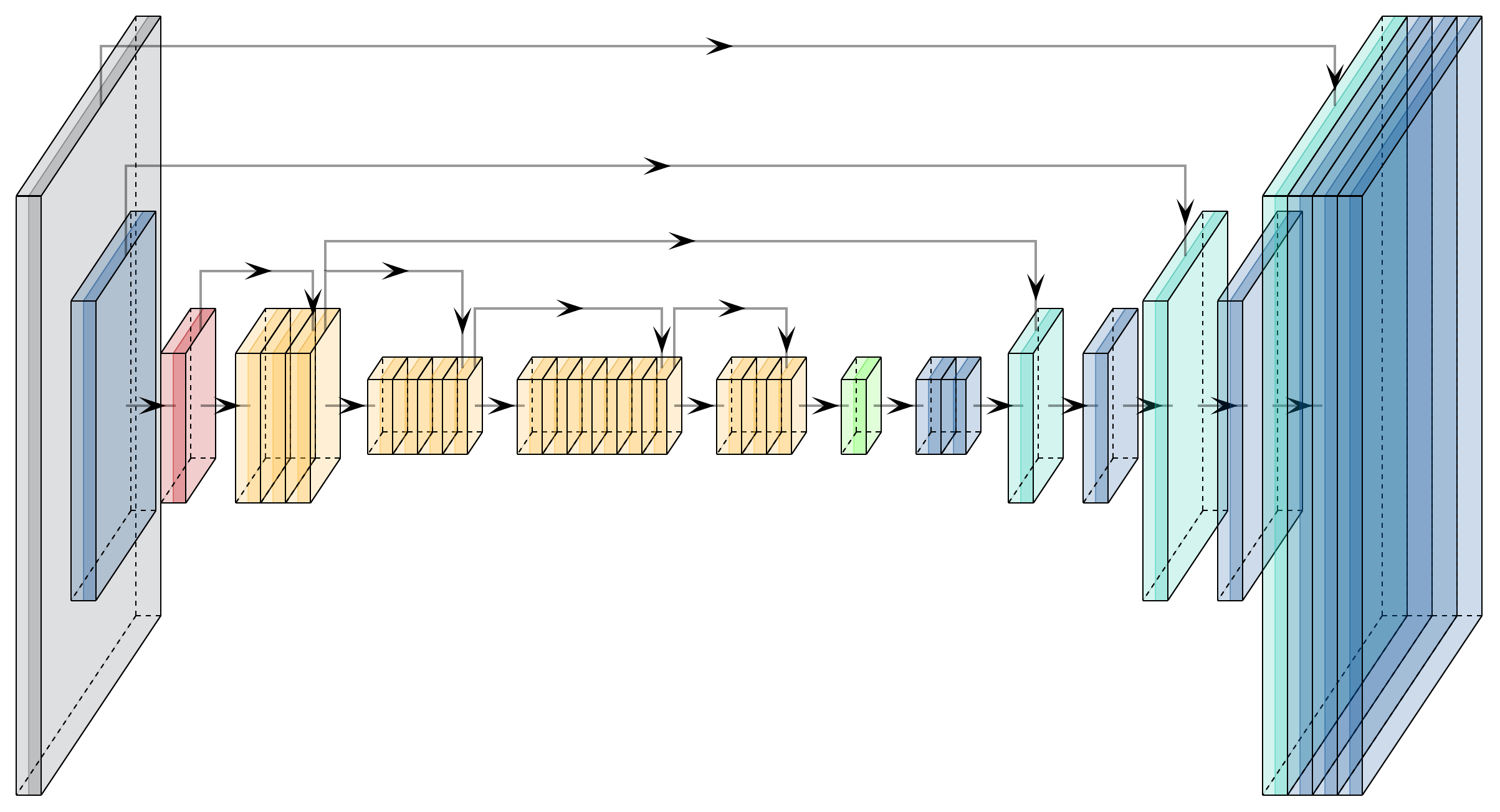}};
	\node at (-0.8, -2.0) {
			\small
			\begin{tcolorbox}[width=5cm, boxsep=5pt, boxrule=0.15mm, colback=white,left=0pt,right=0pt,top=0pt,bottom=0pt]
			\textcolor[HTML]{5A5F63}{\rule{2.5mm}{2.5mm}} input\\
			\textcolor[HTML]{044B94}{\rule{2.5mm}{2.5mm}} (strided) convolution\\
			\textcolor[HTML]{BB0308}{\rule{2.5mm}{2.5mm}} max pool\\
			\textcolor[HTML]{FFb62E}{\rule{2.5mm}{2.5mm}} (strided) bottleneck \cite{he2016deep}\\
			\textcolor[HTML]{63FF40}{\rule{2.5mm}{2.5mm}} pyramid pooling \cite{zhao2017pyramid}\\
			\textcolor[HTML]{22c7b1}{\rule{2.5mm}{2.5mm}} upscaling
			\end{tcolorbox}
	};
    \end{tikzpicture}
    \caption{Feature maps of FBA-Net \cite{forte2020fba}. The output of the initial strided convolution (blue) of the input image (gray) is transformed with a max-pool layer (red), followed by 16 bottleneck layers (yellow), one pyramid pooling layer (green) and a mix of convolutions (blue) and upsampling operations (turquoise). The skip connections are indicated with arrows.}
    \label{fig:architecture}
\end{figure}

When applying deep learning to the deblurring problem, many model architectures are possible. A popular choice for image-based tasks is the U-Net architecture \cite{ronneberger2015u}. This architecture is characterized by its encoder-decoder structure to aggregate information across large image regions as well as skip connections to preserve previously learned features across network layers.
We evaluated several U-Net architectures. In the end, we have chosen the model by Forte and Pitié~\cite{forte2020fba}, which employs a modified ResNet-50 \cite{he2016deep} with increased stride as the encoder in conjunction with weight standardization \cite{qiao2019micro}, group normalization \cite{wu2018group} and a pyramid pooling layer \cite{zhao2017pyramid} (Figure~\ref{fig:architecture}).

\begin{figure}
\centering
\newcommand{\trainsamplewidth}{0.135 \linewidth}
\begin{tabular}{ccccc}
\includegraphics[width=\trainsamplewidth]{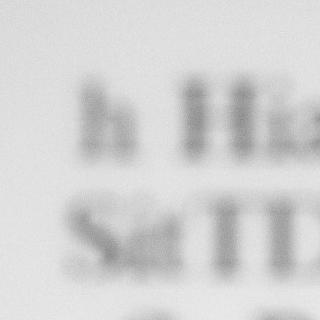} &
\includegraphics[width=\trainsamplewidth]{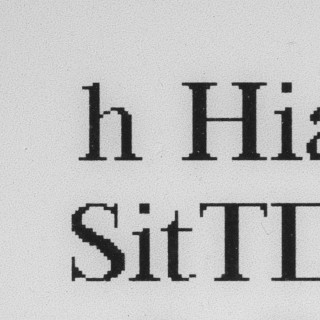} &
\hspace{10mm} &
\includegraphics[width=\trainsamplewidth]{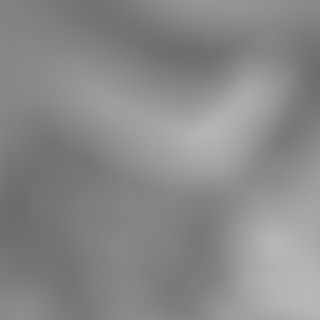} &
\includegraphics[width=\trainsamplewidth]{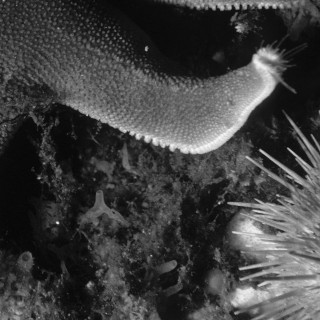}\\
Blurry (HDC2021) & Sharp & & Blurry (DIV2K) & Sharp
\end{tabular}
\caption{Two $320 \times 320$ pairs of cropped training samples from the HDC2021 dataset (left) and DIV2K dataset (right).}
\label{fig:trainingsamples}
\end{figure}

\begin{figure}[htp]
    \centering
    \def\svgwidth{\columnwidth}
\begingroup%
  \makeatletter%
  \providecommand\color[2][]{%
    \errmessage{(Inkscape) Color is used for the text in Inkscape, but the package 'color.sty' is not loaded}%
    \renewcommand\color[2][]{}%
  }%
  \providecommand\transparent[1]{%
    \errmessage{(Inkscape) Transparency is used (non-zero) for the text in Inkscape, but the package 'transparent.sty' is not loaded}%
    \renewcommand\transparent[1]{}%
  }%
  \providecommand\rotatebox[2]{#2}%
  \newcommand*\fsize{\dimexpr\f@size pt\relax}%
  \newcommand*\lineheight[1]{\fontsize{\fsize}{#1\fsize}\selectfont}%
  \ifx\svgwidth\undefined%
    \setlength{\unitlength}{5193.75011534bp}%
    \ifx\svgscale\undefined%
      \relax%
    \else%
      \setlength{\unitlength}{\unitlength * \real{\svgscale}}%
    \fi%
  \else%
    \setlength{\unitlength}{\svgwidth}%
  \fi%
  \global\let\svgwidth\undefined%
  \global\let\svgscale\undefined%
  \makeatother%
  \begin{picture}(1,0.36816968)%
    \lineheight{1}%
    \setlength\tabcolsep{0pt}%
    \put(0,0){\includegraphics[width=\unitlength,page=1]{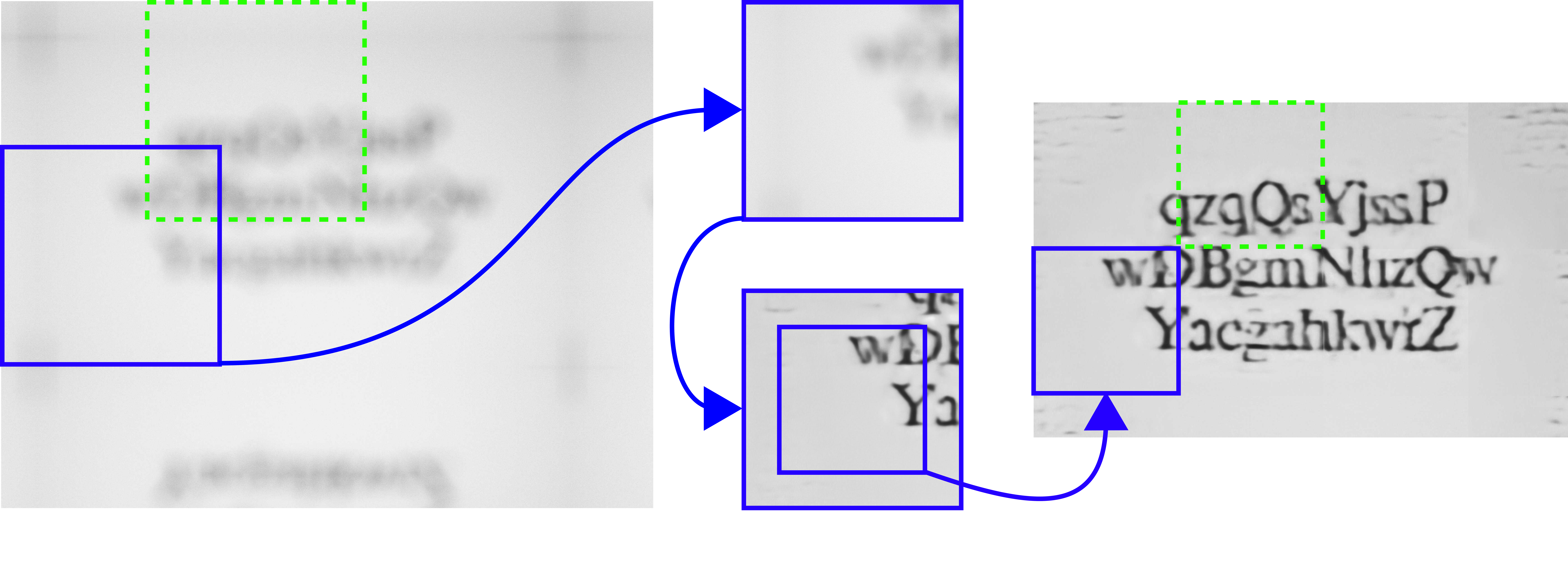}}%
    \put(0.06720795,0.00450559){\makebox(0,0)[lt]{\lineheight{1.25}\smash{\begin{tabular}[t]{l}1. Crop overlapping tile\end{tabular}}}}%
    \put(0.4922608,0.00450559){\makebox(0,0)[lt]{\lineheight{1.25}\smash{\begin{tabular}[t]{l}2. Deblur\end{tabular}}}}%
    \put(0.69167962,0.00450559){\makebox(0,0)[lt]{\lineheight{1.25}\smash{\begin{tabular}[t]{l}3. Crop  Reassemble\end{tabular}}}}%
  \end{picture}%
\endgroup%

    \caption{The padded blurry image is decomposed into overlapping tiles, which are then deblurred, cropped and reassembled (blue). The additional green tile with dashed outline shows that tiles in the blurry input image (left) overlap by 160 pixels, while tiles in the deblurred output image (right) do not.}
    \label{fig:overlapped}
\end{figure}
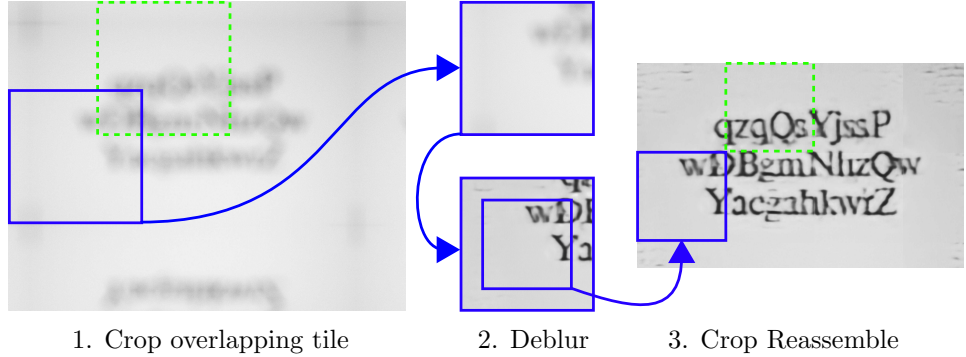

To train the neural network $N$ with parameters $\theta_s$ for stage $s$, we minimize the loss function
\begin{equation}
	\label{eq:deblurloss}
	\mathcal{L}_\text{deblur}(\theta_s) = \sum_{(T^S, T^B) \in D_s} \frob{ T^S - N_{\theta_s}(T^B) }^2
\end{equation}
over sharp and blurry training sample pairs $(T^S, T^B)$ cropped from the dataset $D$.
Unfortunately, optimization over full-sized training images is not possible with our hardware since the images are quite large and training would exceed our memory budget. To work around this issue, we randomly crop $320 \times 320$ image patches from the training sample pairs and train on those instead (Figure~\ref{fig:trainingsamples}). On the one hand, this sets a hard limit on the size of the receptive field of the network, which manifests itself in form of decreasing performance for stages with large blur sizes, but on the other hand, we have two positive side effects. Firstly, it vastly increases the size of the training dataset, thereby reducing the chance of overfitting, and secondly, it decreases the computational cost, which is reflected in a training time of just two hours per stage on an Nvidia RTX 3060 GPU with 6 GB of VRAM.
We employ the Adam optimizer \cite{kingma2014adam} with a learning rate of $10^{-4}$ and train for 50,000 batches with a batch size of 2 and additive Gaussian noise augmentation of scale $3 \cdot 10^{-2}$.
As inference is slightly less costly than training in terms of memory usage, since we do not have to keep track of backpropagation parameters, we can choose a larger cropping size and gain a small increase in accuracy.
Processing an image in a tiled fashion usually introduces boundary artifacts between the patches. Those artifacts can be reduced by processing larger tiles and cropping the overlapped part from the resulting deblurred tiles.
We decompose the blurry images into tiles of size $640 \times 640$ with an additional overlap of $160$ pixels on each side, then deblur the tiles individually, crop the center $640 \times 640$ pixels and reassemble them without overlap into the final image (Figure~\ref{fig:overlapped}).

In order to only process complete tiles, the blurry input image has been reflection padded to the next full tile size. We avoid zero padding or edge padding because when training the neural network later, it would not have any difficulty in recognizing the edge of the image and possibly making different predictions depending on that information. However, we assume a uniform blur, therefore this bias is not desired.

An even more effective strategy to avoid artifacts between tiles is to blend overlapping regions instead of cropping them. The value of pixel $(i, j)$ of the deblurred image $G^D$ can be expressed as a weighted sum of the deblurred tiles $T^D = N_{\theta_s}(T^B)$ with top left corner positioned at $(i_{T^D}, j_{T^D})$ in the reference image
\begin{equation}
G^D_{i,j} = \frac{
\sum_{T^D | (i,j) \in T^D} \, \phi(i - i_{T^D}) \phi(j - j_{T^D}) \, T^B_{i - i_{T^D}, j - j_{T^D}}
}{
\sum_{T^D | (i,j) \in T^D} \, \phi(i - i_{T^D}) \phi(j - j_{T^D}) \hfill
}
\end{equation}
with tile weights
\begin{equation}
\phi(i) =  \left\{
    \begin{array}{ll}
        \frac{1}{2} + \frac{\cos \frac{x}{i}}{2} & \quad i < p \\
        \frac{1}{2} + \frac{\cos \frac{s + 2 p - 1 - x}{i}}{2} & \quad i \ge s + 2 p\\
		1 & \quad \text{otherwise.}
    \end{array}
\right.
\end{equation}
assuming tiles of size $(s + 2 p) \times (s + 2 p)$, overlapping neighboring tiles by $p$ pixels on each side. The sum $\sum_{T^D | (i,j) \in T^D}$ describes the summation over all tiles which contain the pixel $(i, j)$, i.e. tiles which fulfill $i_{T^D} \le i < i_{T^D} + s + 2 p$ and $j_{T^D} \le j < j_{T^D} + s + 2 p$.

\begin{figure}[htp]
		\begin{subfigure}[t]{0.3\textwidth}
			\includegraphics[width=\linewidth]{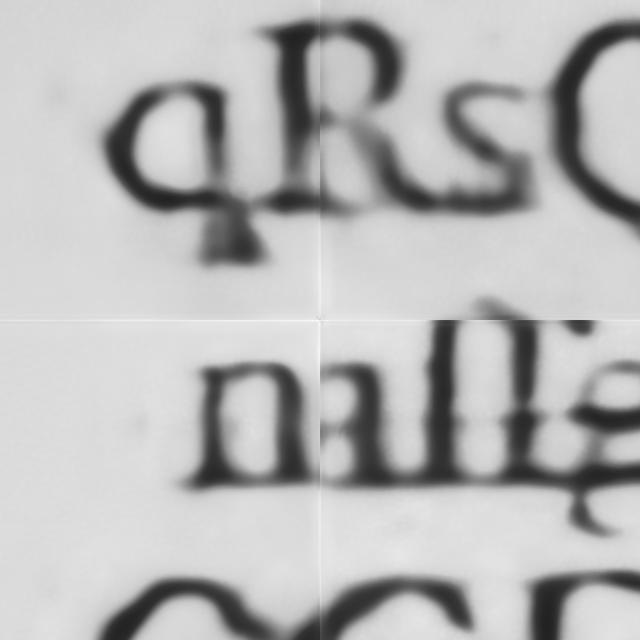}
			\caption{No overlap.}
			\label{fig:tilesnopadding}
		\end{subfigure}
		\hfill
		\begin{subfigure}[t]{0.3\textwidth}
			\includegraphics[width=\linewidth]{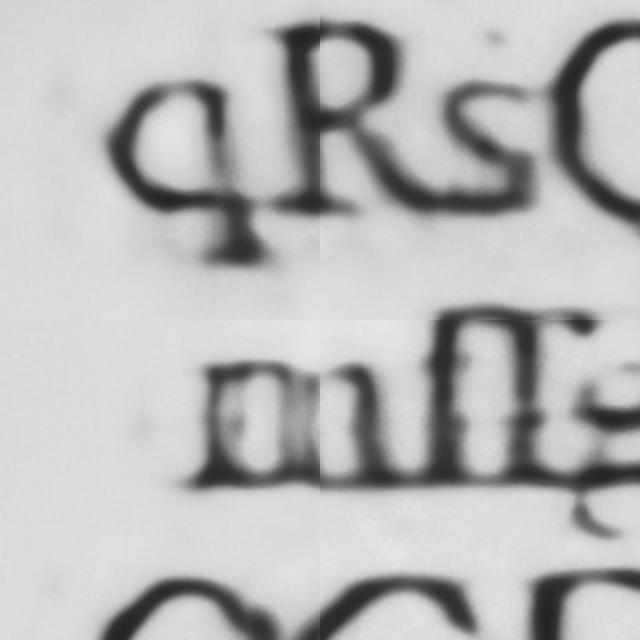}
			\caption{Overlap \& Crop.}
			\label{fig:tilesoverlapped}
		\end{subfigure}
		\hfill
		\begin{subfigure}[t]{0.3\textwidth}
			\includegraphics[width=\linewidth]{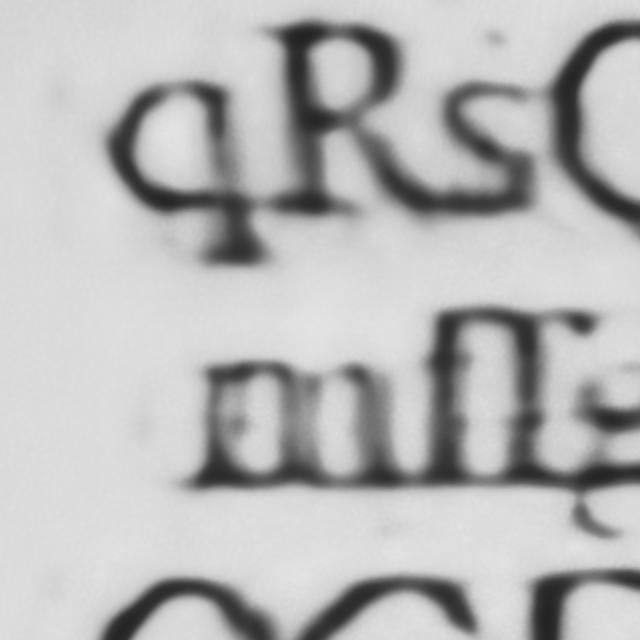}
			\caption{Overlap \& Blend.}
			\label{fig:tilesblended}
		\end{subfigure}
	
	\caption{Comparison between different tile cropping and reassembling methods. A horizontal and vertical seam is clearly visible when neighboring tiles are deblurred individually. Cropping overlapped tiles greatly reduces those artifacts. No apparent seam is visible when blending overlapping tiles.}
	\label{fig:cropvsblend}
\end{figure}

\section{Results}

\begin{figure}
\centering
\newcommand{\cropwidth}{0.16 \linewidth}
\begin{tabular}{ccccc}
\includegraphics[width=\cropwidth]{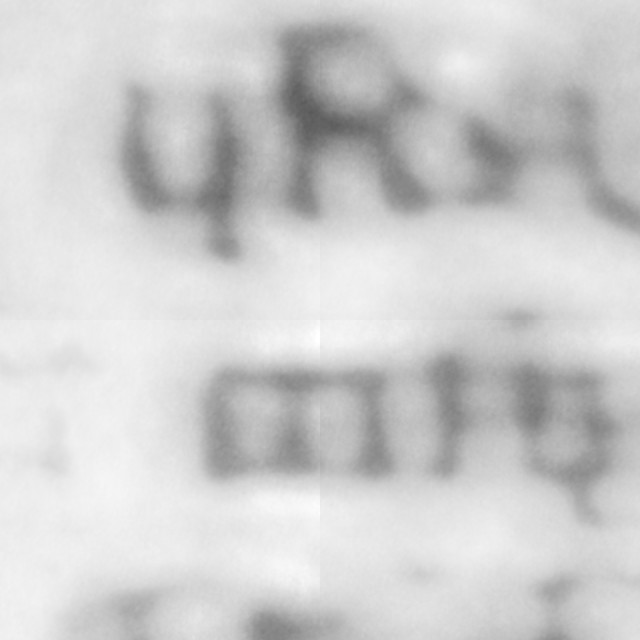} &
\includegraphics[width=\cropwidth]{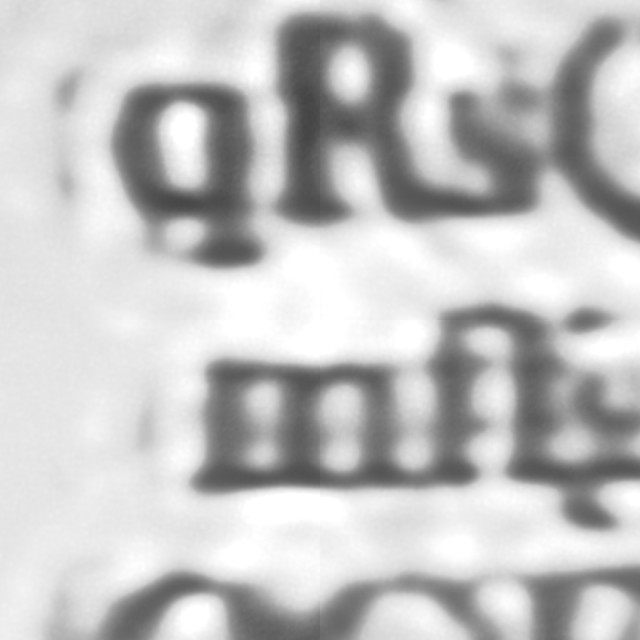} &
\includegraphics[width=\cropwidth]{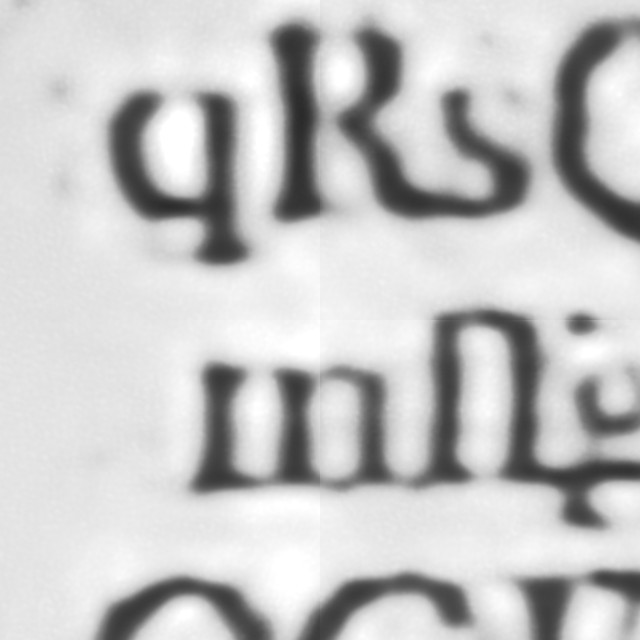} &
\includegraphics[width=\cropwidth]{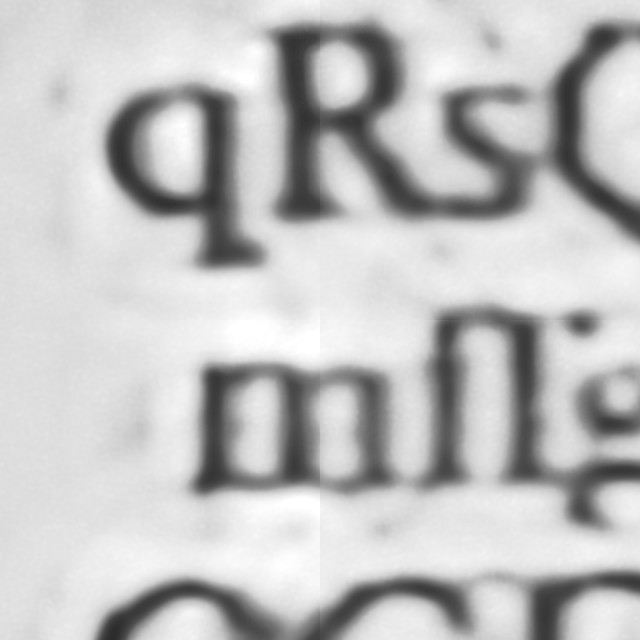} &
\includegraphics[width=\cropwidth]{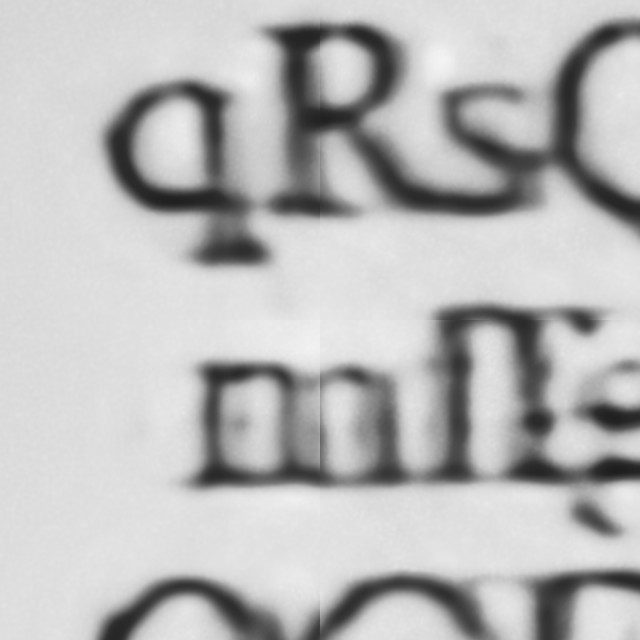}\\
1,000 & 5,000 & 10,000 & 25,000 & 50,000 batches.\\
\end{tabular}
\caption{Evolution of deblurred images from \ourtestdataset{} after training for a certain number of batches.}
\label{fig:training}
\end{figure}

To compare different neural networks and different training methods, we consider the OCR score and the training objective $\mathcal{L}_\text{deblur}(\theta_s)$ (Equation~\ref{eq:deblurloss}) for the last 10 images of stage 9 from \ourtestdataset{}. We fix the crops when calculating $\mathcal{L}_\text{deblur}(\theta_s)$ to get a more meaningful and less noisy result.
\begin{figure}[htp]
    \centering
	\includegraphics[width=\linewidth]{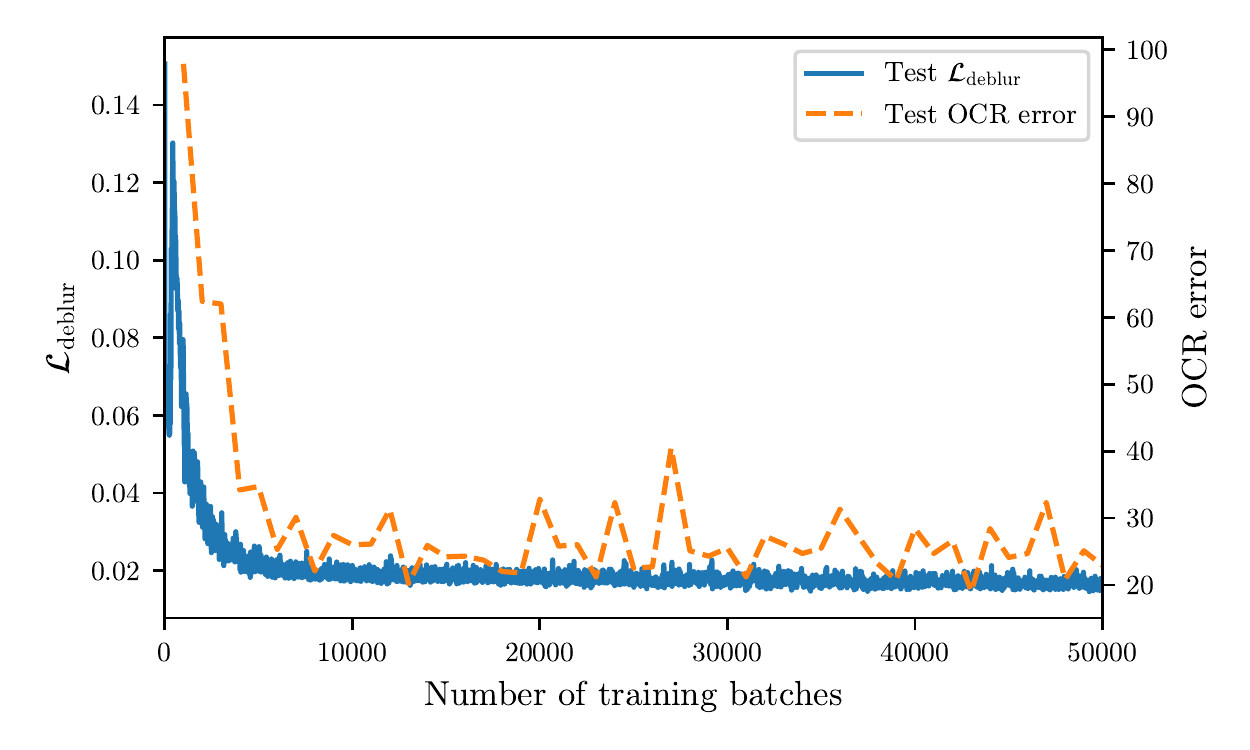}

    \caption{Evolution of $\mathcal{L}_\text{deblur}(\theta_s)$ and the corresponding OCR error for stage 9 of the HDC 2021 dataset over 50000 training batches.}
    \label{fig:lossvsocr}
\end{figure}
In Figure~\ref{fig:lossvsocr}, one can see that $\mathcal{L}_\text{deblur}$ converges relatively quickly after fewer than 10,000 training batches, while the OCR score fluctuates considerably during the entire training procedure. When inspecting the deblurred images visually (Figure~\ref{fig:training}), the image quality improves even after 10,000 batches, suggesting that $\mathcal{L}_\text{deblur}$ is of limited use to evaluate the quality of deblurred images.

When comparing different network architectures
(Table~\ref{tab:networks}), FBA-Net \cite{forte2020fba} performs the
best, followed by IndexNet \cite{lu2019indices}, which still performs
surprisingly well considering that it has 89\% fewer parameters.  In
general, the U-Net architecture \cite{ronneberger2015u} has a worse
performance.  Furthermore, the reconstruction quality is sensitive to
implemention details.
When considering different training methods, it can be seen that augmenting the training batches with Gaussian noise improves the OCR score marginally, while warping the blurry images to align more closely with the sharp images before training improves the score greatly.
\begin{table}
	\centering
	\caption{Comparison of different networks and training methods for stage 9 of the HDC 2021 dataset.}
	\label{tab:networks}
	\begin{tabular}{ | l | r | r | }
		\hline
		Variation & OCR score& Parameters\\
		\hline
		U-Net 1\tablefootnote{\url{https://github.com/mateuszbuda/brain-segmentation-pytorch/blob/master/unet.py}}& 35.85 & $7.76\times 10^6$\\
		U-Net 2\tablefootnote{\url{https://github.com/usuyama/pytorch-unet/blob/master/pytorch_unet.py}} & 4.15 & $7.78\times 10^6$\\
		U-Net 3\tablefootnote{\url{https://github.com/milesial/Pytorch-UNet/blob/master/unet/unet_model.py}} & 3.66 & $31.04\times 10^6$ \\
		IndexNet & 57.38 & $3.69\times 10^6$ \\
		FBA-Net & \textbf{75.34} & $34.67 \times 10^6$\\
		FBA-Net w/o noise aug. & 66.48 & $34.67 \times 10^6$\\
		FBA-Net w/o warping & 29.84 & $34.67 \times 10^6$\\
		\hline
	\end{tabular}
\end{table}
Although more sophisticated cropping strategies produce visually more pleasing results (Figure~\ref{fig:cropvsblend}), their impact on the OCR score is limited (Table~\ref{tab:cropping}).

\begin{table}
	\centering
	\caption{Comparison of different cropping strategies for tiled deblurring of the Times font images of stage 9 of the HDC2021 dataset.}
	\label{tab:cropping}
	\begin{tabular}{ | l | r | }
		\hline
		Overlap & OCR score\\
		\hline
		No overlap & 65.37\\
		Overlap \& crop & 70.27\\
		Overlap \& blend & 71.98\\
		\hline
	\end{tabular}
\end{table}

Our results\footnote{Available at \url{https://www.fips.fi/HDCresults.php\#anchor1}} for the \hdctestdataset{}, which was secret during the challenge, can be seen in Table~\ref{tab:results}. Submissions with OCR scores above a threshold of 70 \% advance to the next stage.
Our OCR scores stay above 95\% up to stage 6, after which the performance degrades gradually, staying barely above the 70\% threshold at stage 10. As we expected those numbers, we did not train additional networks for later stages.
\begin{table}[h]
	\caption{OCR scores for stages 0 to 10 of the HDC2021 dataset}
	\label{tab:results}
	\centering
\resizebox{\textwidth}{!}{%
\begin{tabular}{ | c || c | c | c | c | c | c | c | c | c | c | c | }
\hline
Stage  & 0     & 1     & 2     & 3     & 4     & 5     & 6     & 7     & 8     & 9     & 10 \\
\hline
Score & 96.28 &	95.28 &	95.50 &	96.30 &	96.40 &	97.03 &	94.33 &	91.97 &	85.92 &	73.80 &	70.17\\
\hline
\end{tabular}
}
\end{table}
This performance was sufficient to earn us a rank within the midfield of the competition.
In general, the field was split between methods which used neural networks and
methods which did not, suggesting that neural network based approaches are favorable.
However, it is important to get the training data right.
While we generated additional traning data by applying our forward
model to \textit{natural} images to aid generalization, we did not generate
additional blurred \textit{text} images to avoid overfitting.
Several other competitors with similar methods report
improved performance with this strategy, so we believe that it
might improve the performance of our methods as well.

\section{Future work}

To pass the higher stages of the HDC challenge, larger PSFs become
necessary. This poses new technical challenges because the valid
convolution of images with kernels larger than the image size
complicates the estimation. For the HDC2021 challenge, this is less
consequential as the text region in the images is padded with
sufficient empty space, but with more image content closer to the edge
of the image, it will be more difficult to recover those regions
accurately.
Evaluating the performance on images with content
different from text might provide interesting insights.
Another possibly fruitful direction of research might be the
investigation of more elaborate forward models, for example spatially
varying point spread functions \cite{hirsch2010efficient}.
Lastly, we have shown that cropping is an effective strategy to deblur images
with low memory requirements, but we expect that future hardware generations
will provide larger amounts of VRAM, which will make training on larger image
crops more practical and lead to improved deblurring results.

\section{Conclusion}

In this paper, we presented an approach for image deblurring based on
neural networks and evaluated its performance on the
dataset published during the Helsinki Deblur Challenge 2021.
We compare the performance of several neural networks and arrive at
the conclusion that networks developed for alpha matting are also
well-suited for image deblurring.
Furthermore, we estimate a PSF-based blur model to augment our training
dataset with additional synthetically blurred images to achieve better
generalization performance.
Our method has modest training requirements thanks to a patch-based
approach, but still shows strong deblurring capabilities,
which can make text legible that we were unable to decipher by eye and
thereby exceeds our human capabilities.

\end{document}